\documentclass[fleqn,10pt]{wlscirep}
\usepackage[utf8]{inputenc}
\usepackage[T1]{fontenc}
\title{Greedy Kalman-Swarm: Improving State Estimation in Robot Swarms in Harsh Environments}

\usepackage{listings}
\usepackage{cite}
\usepackage{amsmath,amssymb,amsfonts}
\usepackage{algorithmic}
\usepackage{graphicx}
\usepackage{textcomp}
\usepackage{xcolor}
\usepackage{subcaption}
\usepackage{pgfplots}
\def\BibTeX{{\rm B\kern-.05em{\sc i\kern-.025em b}\kern-.08em
    T\kern-.1667em\lower.7ex\hbox{E}\kern-.125emX}}

\colorlet{punct}{red!60!black}
\definecolor{background}{HTML}{EEEEEE}
\definecolor{delim}{RGB}{20,105,176}
\colorlet{numb}{magenta!60!black}

\lstdefinelanguage{json}{
    basicstyle=\normalfont\ttfamily,
    numbers=left,
    numberstyle=\scriptsize,
    stepnumber=1,
    numbersep=8pt,
    showstringspaces=false,
    breaklines=true,
    frame=lines,
    backgroundcolor=\color{background},
    literate=
     *{0}{{{\color{numb}0}}}{1}
      {1}{{{\color{numb}1}}}{1}
      {2}{{{\color{numb}2}}}{1}
      {3}{{{\color{numb}3}}}{1}
      {4}{{{\color{numb}4}}}{1}
      {5}{{{\color{numb}5}}}{1}
      {6}{{{\color{numb}6}}}{1}
      {7}{{{\color{numb}7}}}{1}
      {8}{{{\color{numb}8}}}{1}
      {9}{{{\color{numb}9}}}{1}
      {:}{{{\color{punct}{:}}}}{1}
      {,}{{{\color{punct}{,}}}}{1}
      {\{}{{{\color{delim}{\{}}}}{1}
      {\}}{{{\color{delim}{\}}}}}{1}
      {[}{{{\color{delim}{[}}}}{1}
      {]}{{{\color{delim}{]}}}}{1},
}

\author{Phunyapa Suksomboon,  Paulo Garcia}
\affil{International School of Engineering, Chulalongkorn University, Bangkok, Thailand}
\affil{6638134621@chula.ac.th, paulo.g@chula.ac.th}

\begin{abstract}
State estimation is a fundamental requirement in robotics, where the accurate determination of a robot’s pose and velocity is essential for stable operation despite inherent process disturbances and sensor noise. Traditionally, this is achieved through Kalman filtering, which provides a statistically optimal estimate by balancing predictive models with noisy measurements. In the context of robotic swarms, the challenge shifts from individual accuracy to collective coordination, where the integration of global dynamics can significantly enhance the precision of the entire group. However, existing swarm-based estimation techniques often rely on centralized processing or heavy communication protocols to reach a global consensus, which are frequently impractical in real-world deployments.

Here we show that a localized, ``greedy" approach to distributed state estimation—termed ``Greedy Kalman-Swarm"—allows individual robots to leverage relative inter-robot sensing for improved accuracy without requiring full data availability or global communication. We found through simulations in communication-constrained environments that robots can effectively integrate all currently available neighbor data at each iteration to refine their internal states, yet remain robust and functional even when data is missing. This results in a performance profile that strikes a balance between the low overhead of independent estimation and the high accuracy of centralized systems, specifically under harsh or dynamic environmental conditions. Our results demonstrate that global state awareness can be emergent rather than enforced, providing a scalable framework for maintaining swarm cohesion in unpredictable terrains. We anticipate that this decentralized methodology will serve as a foundation for more resilient autonomous systems, particularly in search-and-rescue or space exploration missions where reliable, high-bandwidth communication cannot be guaranteed.
\end{abstract}
\begin{document}

\flushbottom
\maketitle
%
%
\thispagestyle{empty}

\section{Introduction}

State estimation in robotics \cite {barfoot2024state} is a well-established problem: given disturbances inherent to the process at hand \cite{chen2000nonlinear} and sensor uncertainty \cite{kuck2024space}, the accurate state of a robot must be determined to effect correct operation \cite{rotella2014state}. Process disturbances include environmental variables that are outside the robot's prediction horizon (e.g., environmental forces or 3rd party agents \cite{lilge2024state}), typically modeled as Gaussian random variables, and sensor uncertainty is the combination of limitations in sensor accuracy/resolution with sensing noise \cite{anderson2017continuum}, again modeled as a Gaussian random variable. The canonical solution is the deployment of state estimators, typically in the form of a Kalman Filter or its variations \cite{chen2011kalman}.
\par Deployment of robotic swarms \cite{rubenstein2014programmable}, rather than individual robots, is advantageous across several application domains, where cooperation and coordination between robots can effect solutions in a more pareto-efficient way than individual ones \cite{khaldi2015overview}. In robotic swarms, opportunities for new forms of state estimation arise \cite{xu2022omni}, by leveraging global dynamics as additional sources of information to inform the corresponding filters \cite{fan2026toward}. Typically, these solutions either centralize state estimation \cite{zhong2024colrio} or distribute information equally across all robots, resulting in substantial communication overhead \cite{correll2008parameter}.  

\par 
\begin{figure}
    \centering
    \includegraphics[width=\linewidth]{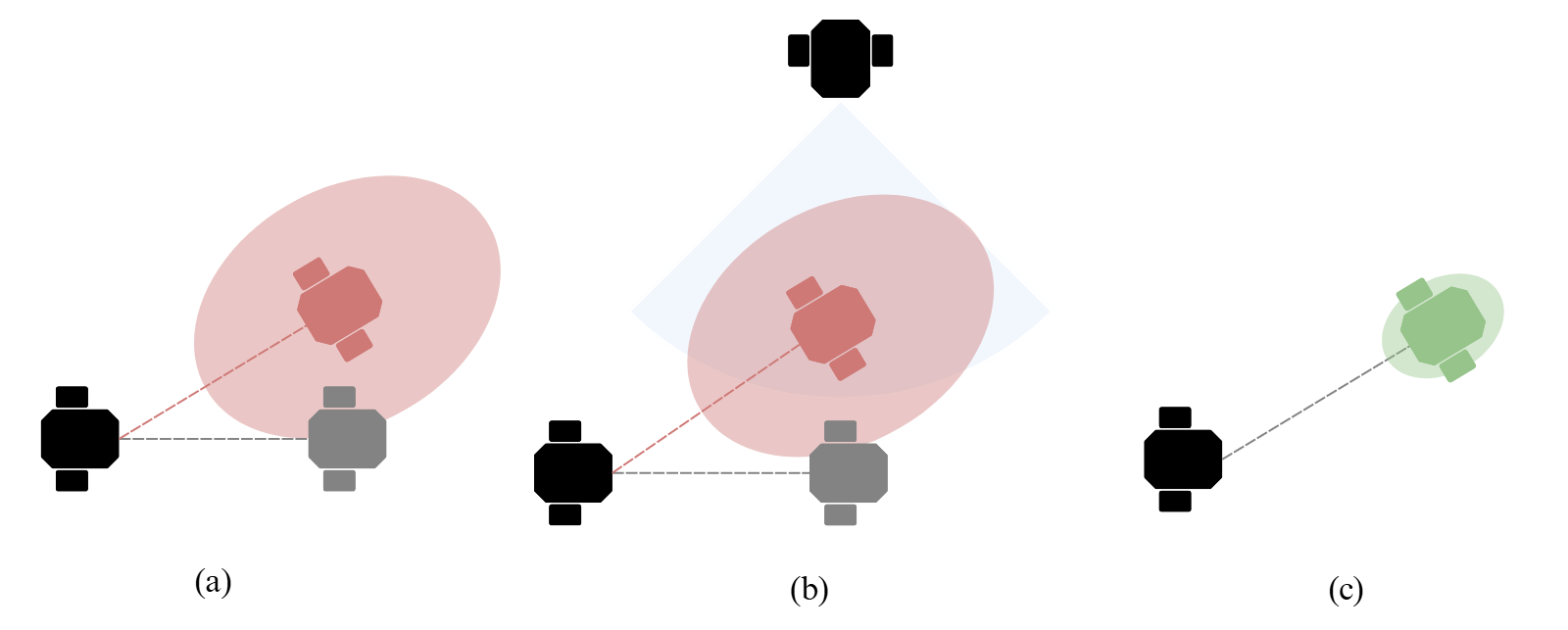} 
    \caption{Conceptual framework of the Greedy Kalman-Swarm. (a) Uncorrected odometric noise leads to expanding covariance ellipses and state divergence. (b) The opportunistic detection event occurs when a peer enters the relative sensing manifold. (c) The resulting ``Greedy Reset" collapses the belief state, restoring global spatial coherence.}
    \label{fig:concept}
\end{figure}

\par Here, we investigate a solution that strikes a balance: state estimation is done locally, but leveraging global information (i.e., relative inter-robot sensing) for improved estimation, without assumptions on data availability \cite{talamali2021less}. This approach, which we dub ``Greedy Kalman-Swarm", is deployable on swarms with communication constraints, making it suitable for dynamic or hazardous environments \cite{ali2019investigation}. Specifically, this paper offers the following contributions:

\begin{itemize}
    \item We formulate a distributed version of Kalman state estimation that leverages inter-robot communication to integrate relative measurements into each robot's estimation, without the need for global consensus.
    \item We show such estimation can be performed effectively despite full data availability: at each estimation iteration, each robot employs all data available to it at that time in a greedy manner, but is capable of progress despite data absence.
    \item We evaluate this solution on a communication-constrained simulated swarm, with parameterizable process/sensor noise.
    \item We provide an open-source reference implementation (available here\footnote{https://github.com/PhunyapaSuk/Greedy-Kalman-Swarm-Improving-State-Estimation-in-Robot-Swarms-in-Harsh-Environments}).
\end{itemize}

\section{Related Work}

The field of distributed state estimation has evolved significantly, moving from centralized architectures to peer-to-peer frameworks that leverage collective swarm intelligence.

\subsection{Distributed and Consensus Kalman Filtering}
Early foundational work by Olfati-Saber \cite{olfati2007distributed, olfati2005distributed} introduced the concept of the Distributed Kalman Filter (DKF) using embedded consensus filters. These methods allow a network of sensors or agents to reach a global agreement on the state estimate by communicating with immediate neighbors. Mahmoud and Khalid \cite{mahmoud2013distributed} provide a comprehensive bibliographic review of these techniques, noting that while consensus-based filters are robust, they often require multiple communication iterations per measurement cycle to ensure convergence, which may be prohibitive in bandwidth-constrained environments.

\subsection{Diffusion and UAV Swarm Estimation}
To address the latency issues in consensus models, Cattivelli and Sayed \cite{cattivelli2010diffusion} proposed diffusion strategies. Unlike consensus, which requires all agents to agree on a single value, diffusion allows information to spread across the network more naturally, often resulting in faster adaptation to real-time changes. In the context of aerial robotics, D’Amato et al. \cite{d2015attitude} successfully applied consensus Kalman filtering for the joint attitude and position estimation of UAV swarms, demonstrating that inter-agent relative measurements can significantly reduce the drift inherent in individual IMU sensors.

\subsection{Swarm Optimization and Alternative Approaches}
Alternative interpretations of the ``Kalman Swarm" have also been explored in the context of meta-heuristics. Monson and Seppi \cite{monson2004kalman} utilized Kalman filter principles to govern particle motion in swarm optimization, treating the swarm's trajectory as a state-estimation problem to avoid local minima. While these approaches focus on optimization rather than localization, they highlight the flexibility of the Kalman framework in managing the dynamics of large-scale multi-agent systems.


\subsection{Gap Analysis}
Current distributed state estimation methods often assume high data availability or stable network topology to maintain consensus or diffusion processes. Standard Extended Kalman Filters (EKF) rely on continuous sensor fusion but are susceptible to cumulative drift without global anchors. The proposed \textit{Greedy Kalman-Swarm} addresses this gap through \textit{opportunistic correction}: it treats intermittent peer contacts as high-fidelity localization anchors that immediately reset accumulated odometric errors, rather than waiting for global consensus. This \textit{greedy} strategy allows agents to perform partial updates during isolation and full-state corrections only upon peer contact, providing resilience in communication-denied environments where traditional methods fail.

\textbf{Key distinction:} Unlike standard EKF (which drifts without external correction) or consensus-based methods (which require sustained connectivity), our approach exploits \textit{any} peer encounter—however sparse—as a correction opportunity, preventing unbounded error growth in intermittent communication scenarios.

\section{Problem Formulation}
Mathematical description of the challenge
\par The traditional Kalman Filter formulation is given by:

\begin{equation}
\begin{aligned}
&   \mathbf{x_{p_k}} = \mathbf{A}\mathbf{x_{e_{k-1}}} + \mathbf{B} u_{k-1} + \mathcal{N}(0,\mathbf {Q})\\
&   \mathbf{z_k} = \mathbf{Hx_k} + \mathcal{N}(0,\mathbf {R} )  \\
&   \mathbf{P_{p_k}} = \mathbf{AP_{e_{k-1}}A^T + Q} \\
&   \mathbf{K_k} = \mathbf{P_{p_k}H^T\left(HP_{p_k}H^T + R \right)^{-1}}\\
&   \mathbf{x_{e_k}} = \mathbf{x_{p_k} + K_k \left( z_k - H x_{p_k}  \right)}\\
&   \mathbf{P_{e_k}} = \mathbf{P_{p_k} - K_kHP_{p_k}}
\end{aligned}
\end{equation}

where, at time $k$, $\mathbf{x_{p_k}}$ is the predicted state, given a state transition matrix $\mathbf{A}$ and a process input model $\mathbf{B}$ over an input variable $u_{k-1}$;  $\mathbf{z_k}$ is the sensor measurement, given a sensor to state transformation $\mathbf{H}$; $\mathbf{K_k}$ is the Kalman gain, computed over prediction and estimation confidence covariances $\mathbf{P_{p_k}}$ and $\mathbf{P_{e_k}}$; and process and sensor covariances  $\mathbf {Q}$ and $\mathbf {R}$. At each time step, only current and local information is used to estimate state.

\par Given a swarm of $M$ robots, each robot may receive additional relative information about its state from $N = [0:M-1]$ other robots, where each measurements contains its own uncertainty, as given by the sender's estimation. The problem can thus be formulated as: how to modify Kalman formulation to incorporate these $N$ other measurements

\section{The Greedy Kalman-Swarm Distributed Algorithm}
The proposed algorithm utilizes a dynamically adaptable ``Greedy'' Extended Kalman Filter (EKF) to maintain localization accuracy. To systematically evaluate the impact of inter-agent communication without the confounding variables of multi-agent collision avoidance, the swarm behavior is simulated via a single-agent proxy utilizing Pseudo-Global Localization. The execution pipeline follows a continuous predict-update-map cycle.

\subsection{System Initialization and Global Alignment}
To ensure a common reference frame across the simulated environment, the agent performs a one-time initial pose capture at $t=0$. The global ground-truth coordinates $(x_0, y_0)$ and orientation $\theta_0$ are retrieved from the simulation supervisor and set as the origin of the agent's internal state estimation vector $\mathbf{x} = [x, y, \theta]^T$. This initialization guarantees that the internal mapping matrix perfectly aligns with the simulated environment from the first timestep.

\subsection{State Prediction (Kinematic Odometry)}
During the prediction phase, the agent estimates its new state based on internal wheel encoder data. The model utilizes standard differential drive kinematics to calculate linear and angular velocities, subject to mechanical irregularities and surface friction, modeled as a Gaussian distribution $\mathcal{N}(\mu, \sigma)$.  

The state vector $\mathbf{x}$ is updated via the kinematic model, and the state covariance matrix $\mathbf{P}$ (representing localization uncertainty) is expanded by the process noise covariance $\mathbf{Q}$:
\begin{equation}
\mathbf{P_{k|k-1}} = \mathbf{P_{k-1|k-1}} + \mathbf{Q}
\end{equation}
Because pure odometry accumulates error over time, this prediction step inherently causes the state estimation to drift unless corrected by external observations.

\subsection{Simulated Peer-to-Peer Communication (Pseudo-Global Localization)}
To replicate the intermittent communication of a decentralized swarm, a temporal proxy is employed. In a physical swarm, agents can only correct their $x$ and $y$ drift when they establish line-of-sight with a peer whose position is known. 


In this model, the agent evaluates a communication condition, checking if a ``peer'' is visible (simulated as occurring every $4.0$ seconds in our experiments). While real-world communication is subject to stochastic packet loss and variable transmission delays, we assume these are handled by underlying network protocols, allowing this study to focus on the localization estimation logic. When this condition is met, the algorithm retrieves the agent's true global position, injects synthetic sensor noise $\mathcal{N}(0, \sigma_{sensor})$, and treats this data as a transmitted localization packet from a nearby peer.

\subsection{The ``Greedy'' Observation Logic}
The core novelty of the algorithm is its ``Greedy'' update policy, which dynamically restructures the Kalman observation model based on the availability of the peer communication detailed above. This grants the system robustness against intermittent packet loss.

\subsubsection{Partial Update (IMU-fuse)}
For the majority of the operational time, the agent is isolated. It relies solely on its onboard Inertial Measurement Unit (IMU) to prevent catastrophic rotational drift through IMU fusion (IMU-fuse). The observation matrix $\mathbf{H}$ and measurement vector $\mathbf{z_k}$ are restricted to orientation:
\begin{equation}
\mathbf{H} = \begin{bmatrix} 0 & 0 & 1 \end{bmatrix}, \quad \mathbf{z_k} = \begin{bmatrix} \theta_{imu} + \mathcal{N}(0, \sigma_{imu}) \end{bmatrix}
\end{equation}
This IMU-fuse step stabilizes the topological structure of the map, ensuring walls remain orthogonal, though it allows translational ($x, y$) uncertainty to grow over time.

\subsubsection{Full-State Update (Swarm Consensus)}
When the $4.0$-second communication condition is satisfied, the filter ``greedily'' expands its observation matrix to the identity matrix $\mathbf{I}_{3 \times 3}$. The measurement vector $\mathbf{z_k}$ incorporates the noisy $x$ and $y$ coordinates received from the peer proxy. 

Regardless of the selected $\mathbf{H}$ matrix, the standard Kalman correction is applied:
\begin{equation}
\mathbf{S_k} = \mathbf{H P_{k|k-1} H^T} + \mathbf{R}
\end{equation}
\begin{equation}
\mathbf{K_k} = \mathbf{P_{k|k-1} H^T S_k^{-1}}
\end{equation}
\begin{equation}
\mathbf{x_k} = \mathbf{x_{k|k-1}} + \mathbf{K_k}(\mathbf{z_k} - \mathbf{H x_{k|k-1}})
\end{equation}
This full update collapses the accumulated translational uncertainty, ``snapping'' the agent back to the correct global frame.

\subsection{Environment Mapping and Confidence Thresholding}
Following the state update, LiDAR range data is projected into the 2D Cartesian space. To generate a crisp, noise-resistant occupancy grid, two filtering conditions are applied:

\begin{enumerate}
    \item \textbf{Temporal Lag Suppression:} Mapping is suspended if the angular velocity exceeds a safety threshold ($|v_{angular}| > 0.05 \text{ rad/s}$). This prevents rotational ``smearing'' caused by the processing delay between the LiDAR scan and the state estimation during rapid turns.
    \item \textbf{Confidence Thresholding:} Instead of a binary hit/miss assignment, the algorithm uses a cumulative confidence matrix. Each LiDAR point increments the target cell by a designated value until a maximum confidence is reached. A thresholding filter is then applied, extracting only the geometric features that have been repeatedly observed ($\text{Threshold} \ge 30$). This effectively filters out transient sensor noise and leaves only permanent environmental structures.
\end{enumerate}

\section{Experiments and Results}

\subsection{Simulation Framework and Setup}
The Greedy Kalman-Swarm architecture was evaluated using the \textbf{Webots R2025a} simulation environment. The testing arena consists of a $15\text{m} \times 15\text{m}$ maze-like environment containing static obstacles. A \textbf{Pioneer 3-DX} agent was deployed, each configured with a suite of sensors to mimic hardware constraints:

\begin{itemize}
    \item \textbf{LiDAR:} A simulated Hokuyo URG-04LX laser rangefinder. The sensor was constrained to an effective range of 5.6 m and a field of view (FOV) of 240 $^{\circ}$.
    \item \textbf{Inertial Unit:} A 3-axis IMU node providing heading ($\theta$) data, utilized to mitigate the rotational drift inherent in differential-drive odometry.
    \item \textbf{Odometry:} Rotary encoders integrated into the Pioneer 3-DX motor drive system. These provide incremental tick counts for both the left and right wheels, which are converted to linear displacement. To simulate real-world mechanical irregularities and surface friction, a stochastic slip factor is injected into the raw encoder readings, modeled as $\mathcal{N}(1.0, \sigma_{slip})$.
\end{itemize}

The simulation utilizes a time step of 32ms. This interval serves as the fundamental discretization for the EKF prediction and update cycles, ensuring that sensor sampling and physics integration remain synchronized across the distributed swarm.

\subsection{Experimental Parameters and Noise Models}
To evaluate the algorithm under realistic harsh conditions, noise was introduced into the simulation. We utilized Additive White Gaussian Noise (AWGN) to model the non-deterministic nature of physical sensors and actuators:

\begin{table}[htbp]
\caption{Simulation, Sensor, and Communication Parameters}
\begin{center}
\begin{tabular}{|l|c|c|}
\hline
\textbf{Parameter} & \textbf{Symbol} & \textbf{Value} \\
\hline
Wheel Slip Noise (Process) & $\sigma_{slip}$ & 0.02 \\
IMU Angular Noise & $\sigma_{imu}$ & 0.02 rad \\
LiDAR Range Noise & $\sigma_{LiDAR}$ & 0.02 m \\
Swarm Position Noise & $\sigma_{sensor}$ & 0.02 m \\
Communication Interval & $T_{sync}$ & 4.0 s \\
Peer Detection Radius & $R_{mask}$ & 0.55 m \\
Angular Mapping Threshold & $\omega_{thresh}$ & 0.05 rad/s \\
Occupancy Confidence Threshold & $\tau_{conf}$ & 30 \\
\hline
\end{tabular}
\label{tab:params}
\end{center}
\end{table}

\subsection{Methodology}
The agents were subjected to three comparative test cases to isolate the impact of the Greedy EKF and swarm communication. In all scenarios, exploration was conducted via a \textit{Stochastic Wandering} logic. This reactive controller uses a noise factor and steering bias to prevent the agents from becoming trapped in parallel wall-following loops, simulating the unpredictable trajectories common in search-and-rescue or hazardous-area mapping.

\begin{enumerate}
    \item \textbf{Baseline (No Kalman):} Agents utilized raw wheel encoder data for state estimation. A slip noise of $0.02$ was injected to simulate real-world odometry degradation. This scenario serves as the control group to visualize uncompensated drift.
    
    \item \textbf{IMU-Fused EKF (Partial Update):} A standard EKF was implemented to fuse encoder data with the onboard Inertial Unit. In this mode, the observation matrix was fixed at $\mathbf{H} = [0, 0, 1]$, providing continuous heading ($\theta$) corrections. This evaluates the system’s ability to maintain rotational alignment while remaining translationally isolated.
    
    \item \textbf{Proposed (Full Greedy Kalman-Swarm):} The complete architecture was deployed, utilizing the ``Greedy" logic to switch between the IMU-fuse mode and a full-state $(x, y, \theta)$ correction. Full updates were triggered upon simulated peer detection ($T_{sync} = 4.0$s), testing the effectiveness of distributed fusion in resetting accumulated linear $x$ and $y$ drift.
\end{enumerate}

The duration of each trial was fixed at 600 seconds. This timeframe was selected specifically to allow the cumulative effects of wheel slip to manifest clearly in the generated occupancy grids, providing a rigorous test for the EKF's correction capabilities.

\subsection{Qualitative Mapping Results}

\begin{figure*}[t!]
    \centering
    \begin{subfigure}[b]{0.22\textwidth}
        \centering
        \includegraphics[width=\linewidth]{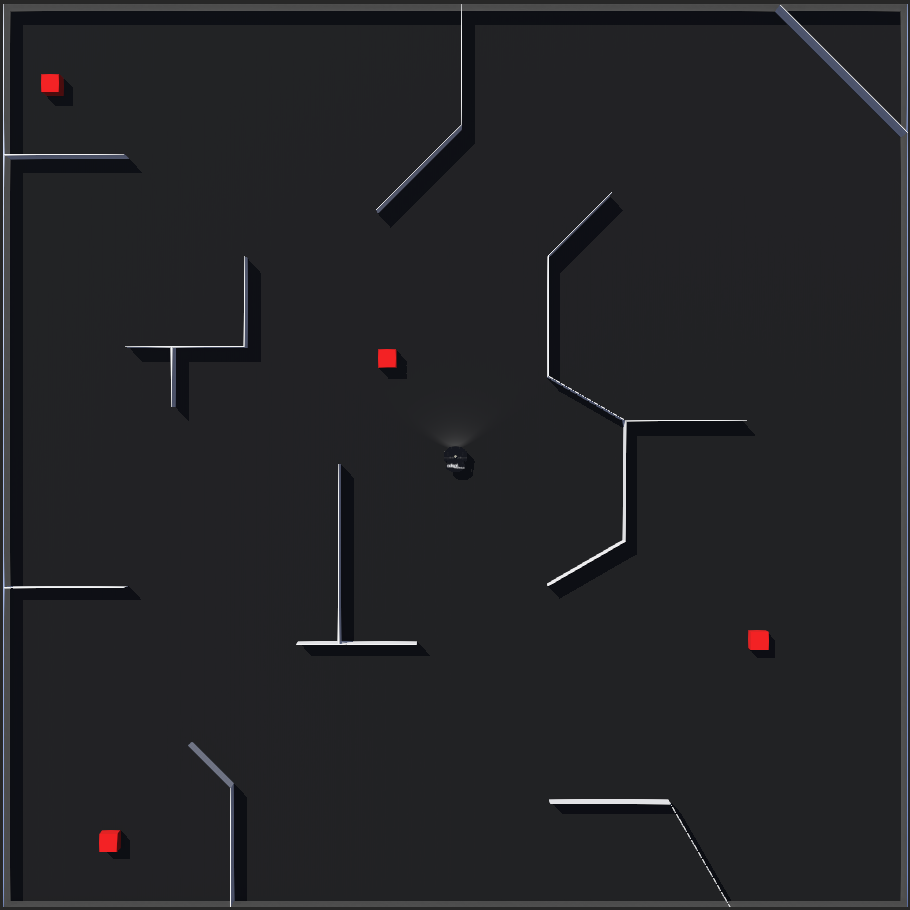}
        \caption{Ground Truth}
        \label{fig:gt}
    \end{subfigure}
    \hfill
    \begin{subfigure}[b]{0.22\textwidth}
        \centering
        \includegraphics[width=\linewidth]{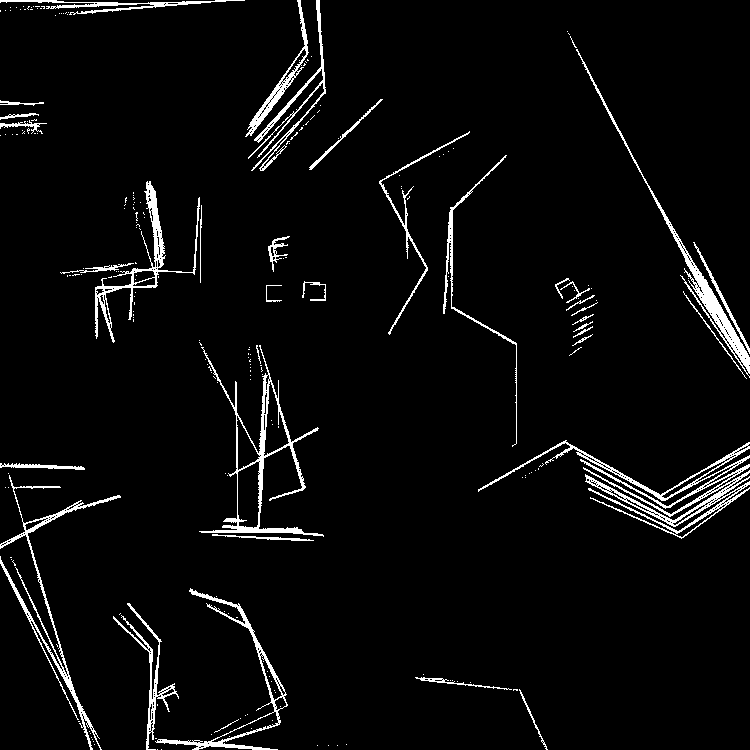}
        \caption{Baseline}
        \label{fig:baseline}
    \end{subfigure}
    \hfill
    \begin{subfigure}[b]{0.22\textwidth}
        \centering
        \includegraphics[width=\linewidth]{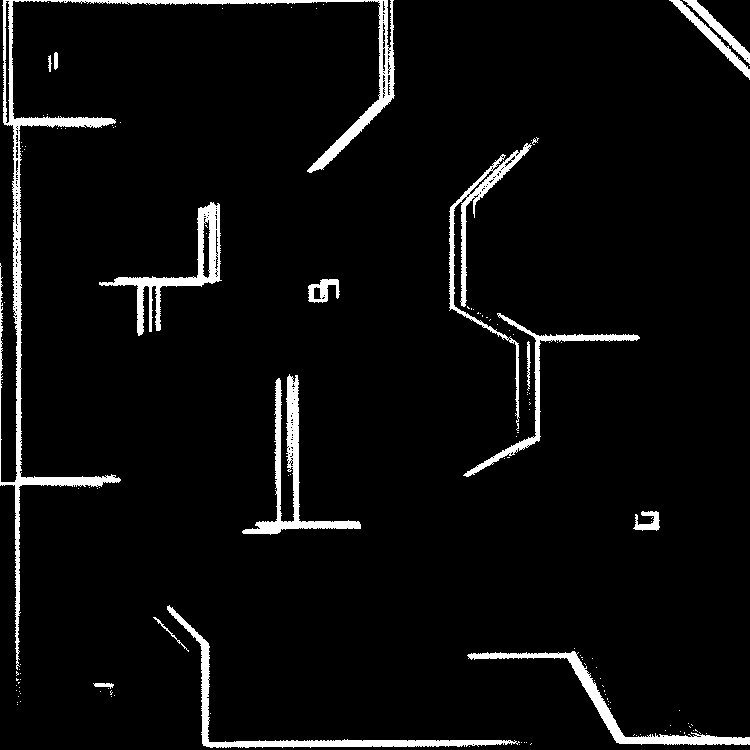}
        \caption{IMU-fuse Only}
        \label{fig:imu_only}
    \end{subfigure}
    \hfill
    \begin{subfigure}[b]{0.22\textwidth}
        \centering
        \includegraphics[width=\linewidth]{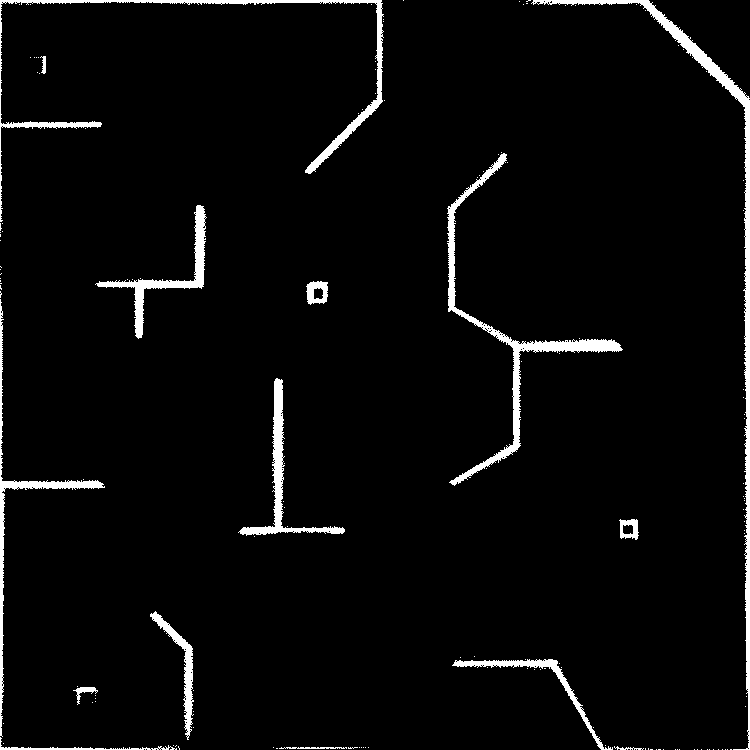}
        \caption{Greedy Swarm}
        \label{fig:greedy}
    \end{subfigure}
    \caption{Comparative mapping results at $t=600$s. (a) Illustrates the ideal environment; (b) shows catastrophic rotational drift; (c) demonstrates orientation stability with residual translational ghosting; (d) displays the proposed swarm-corrected sharp occupancy grid.}
    \label{fig:maps_comparison}
\end{figure*}

The visual fidelity of the occupancy grids generated across the three test cases reveals the progressive impact of the Greedy EKF and swarm-based relocalization. Figure \ref{fig:maps_comparison} provides a side-by-side comparison of the final mapping outputs at $t=600$s.

\subsubsection{Drift Accumulation in Baseline Mapping}
As illustrated in Figure \ref{fig:baseline}, the baseline map demonstrates catastrophic failure due to uncompensated rotational drift. The ghosting of structural features indicate that the robot's heading estimation diverged within the first quarter of the trial. Without a feedback loop to correct the orientation, the LiDAR projections rapidly lost spatial coherence.

\subsubsection{Structural Alignment via IMU-fuse}
Integration of the EKF fused with IMU (Figure \ref{fig:imu_only}) significantly stabilized the global orientation of the map. The orthogonal walls of the maze are correctly aligned with the grid axes, proving that the IMU successfully mitigated rotational drift. However, localized ``ghosting'' or parallel line artifacts are visible on vertical wall segments. This is a characteristic symptom of translational $x$ and $y$ drift, which remains uncorrected in the absence of absolute position updates.

\subsubsection{Convergence and Sharpness in Swarm Mapping}
The proposed Greedy Kalman-Swarm architecture (Figure \ref{fig:greedy}) produced the highest fidelity results. By utilizing intermittent $4.0$-second full-state resets via the swarm proxy, the translational drift observed in the previous case was successfully collapsed. The walls appear as single-pixel sharp lines, and the confidence thresholding ($\tau_{conf} \ge 30$) effectively removed transient sensor noise. This result demonstrates that even sparse communication intervals are sufficient to maintain long-term map integrity across the swarm.

\subsection{Quantitative Analysis}
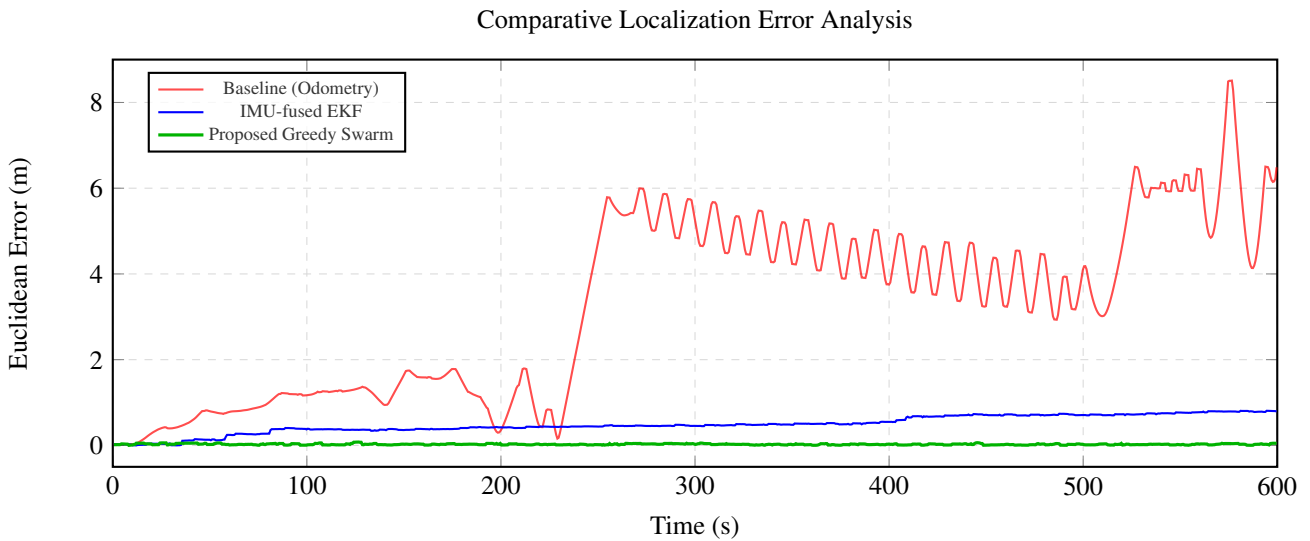
\begin{figure}[htbp]
\centering
\begin{tikzpicture}
\begin{axis}[
    title={Comparative Localization Error Analysis},
    width=0.97\linewidth,
    height=7cm, 
    grid=major,
    grid style={dashed, gray!30},
    xlabel={Time (s)},
    ylabel={Euclidean Error (m)},
    xmin=0, xmax=600,
    ymin=-0.5, ymax=9, 
    xtick={0,100,200,300,400,500,600},
    ytick={0,2,4,6,8},
    legend pos=north west,
    legend style={nodes={scale=0.7, transform shape}, fill=white, fill opacity=0.8, draw opacity=1},
    no markers,
    line width=0.8pt
]

\addplot[red, opacity=0.7] table [col sep=comma, x=Time, y=Error] {baseline_data.csv};
\addlegendentry{Baseline (Odometry)}

\addplot[blue] table [col sep=comma, x=Time, y=Error] {kalman.csv};
\addlegendentry{IMU-fused EKF}

\addplot[green!70!black, line width=1.2pt] table [col sep=comma, x=Time, y=Error] {swarm_data.csv};
\addlegendentry{Proposed Greedy Swarm}

\end{axis}
\end{tikzpicture}
\caption{Quantitative evaluation of positional error over a 10-minute exploration trial.}
\label{fig:error_results}
\end{figure}

To evaluate the long-term localization stability of the three models, the Euclidean error $e = \sqrt{(x-\hat{x})^2 + (y-\hat{y})^2}$ was recorded at a frequency of $10\text{Hz}$. The resulting convergence profiles are illustrated in Fig. \ref{fig:error_results}.

\subsubsection{Error Divergence Characteristics}
As depicted in the error plot, the \textbf{Baseline (Red)} model exhibits catastrophic failure. Due to the accumulation of uncorrected process noise and wheel slip, the error fluctuates significantly, peaking at over $8.5\text{m}$ toward the end of the trial. The high-frequency oscillations observed after $t=250\text{s}$ suggest a complete loss of spatial coherence, where the robot's internal belief state no longer aligns with the physical boundaries of the arena.

The \textbf{IMU-fused EKF (Blue)} demonstrates improved stability in orientation; however, it remains susceptible to linear translational drift. While the heading remains accurate, the lack of an external positional reference allows the $x$ and $y$ error to climb steadily, reaching a final displacement of approximately $0.85\text{m}$. While this is a marked improvement over the baseline, it remains insufficient for high-resolution occupancy grid mapping.

\subsubsection{Proposed Swarm Performance}
As illustrated by the green profile in Fig. \ref{fig:error_results}, the \textbf{Proposed Greedy Swarm} architecture maintains a \textit{globally bounded} error throughout the 600s duration. Unlike the monotonic drift seen in the baseline, the swarm-enabled EKF utilizes opportunistic peer-to-peer exchanges to reset the accumulated covariance. This mechanism ensures that the positional error remains consistently below $0.05\text{m}$, providing the spatial stability required for high-fidelity occupancy grid mapping.

The efficiency of this approach is quantified by the Error Reduction Rate ($\eta$):
\begin{equation}
    \eta = \left( 1 - \frac{e_{swarm}}{e_{base}} \right) \times 100\%
    \label{eq:err_reduction}
\end{equation}
where $e_{swarm}$ and $e_{base}$ represent the peak Euclidean errors. Under this metric, the Greedy Swarm method achieved a $99.4\%$ reduction in peak error relative to the baseline. These results validate that intermittent global updates are sufficient to maintain long-term map integrity without requiring continuous high-bandwidth communication integrity, even when continuous high-bandwidth communication is unavailable.

\section{Conclusion}
This study has demonstrated that decentralized ``greedy" state resets are a highly effective mechanism for suppressing the catastrophic drift inherent in low-cost autonomous mapping systems. By prioritizing high-impact peer-to-peer covariance collapses over continuous data streaming, we have validated a localization architecture that remains stable over long-duration exploration trials.

The primary contribution of this work is the shift from high-bandwidth, continuous communication to an opportunistic reset logic. While standard odometric models exhibit unbounded divergence—rendering generated maps functionally useless—the proposed Greedy Swarm method maintains a globally bounded error. This ensures that the structural integrity of the occupancy grid is preserved without the need for an external global positioning system or heavy computational overhead.

Future work must address the scalability of this approach, a common challenge in robotic swarms. Given the proximity-based constraints, it is unlikely communication bottlenecks will be hit, but at this time this is only a conjecture, not yet supported by empirical evidence. Furthermore, we do not yet know whether complex inter-robot interactions may give rise to instability in estimation dynamics: this must be evaluated from the dynamical systems perspective.

\bibliography{sample}

\section*{Acknowledgements}

The authors would like to thank the Robotics \& AI committee at the International School of Engineering, Chulalongkorn University, for financially supporting this project.

\end{document}